\pdfoutput=1
\documentclass[11pt]{article}

\usepackage[preprint]{acl}

\usepackage{times}
\usepackage{latexsym}
\usepackage{multirow}
\usepackage{booktabs}
\usepackage{adjustbox}

\usepackage{array}
\usepackage[T1]{fontenc}
\usepackage[utf8]{inputenc}

\usepackage{microtype}

\usepackage{inconsolata}

\usepackage{graphicx}
\usepackage{algorithm}
\usepackage{algorithmic}
\usepackage{hyperref}
\usepackage{listings}
\usepackage{fancyvrb}
\usepackage{amssymb}
\usepackage{xcolor}
\usepackage{pifont}
\usepackage{svg}
\usepackage{amsmath}

\usepackage{tcolorbox}
\usepackage{lipsum}
\usepackage{chngcntr}
\title{\textsc{BnSentMix:} A Diverse Bengali-English Code-Mixed Dataset\\ for Sentiment Analysis}

\author{Sadia Alam, Md Farhan Ishmam, Navid Hasin Alvee,\\
\textbf{Md Shahnewaz Siddique, Abu Raihan Mostofa Kamal, Md Azam Hossain}
 \\
  Department of Computer Science and Engineering, Islamic University of Technology\\
  \texttt{\{sadiaalam,farhanishmam,navidhasin,shahnewaz,raihan.kamal,azam\}@iut-dhaka.edu} \\}

\begin{document}
\maketitle
\begin{abstract}
The widespread availability of code-mixed data in digital spaces can provide valuable insights into low-resource languages like Bengali, which have limited annotated corpora. Sentiment analysis, a pivotal text classification task, has been explored across multiple languages, yet code-mixed Bengali remains underrepresented with no large-scale, diverse benchmark. Code-mixed text is particularly challenging as it requires the understanding of multiple languages and their interaction in the same text. We address this limitation by introducing \textsc{BnSentMix}, a sentiment analysis dataset on code-mixed Bengali comprising 20,000 samples with $4$ sentiment labels, sourced from Facebook, YouTube, and e-commerce sites. By aggregating multiple sources, we ensure linguistic diversity reflecting realistic code-mixed scenarios. We implement a novel automated text filtering pipeline using fine-tuned language models to detect code-mixed samples and expand code-mixed text corpora. We further propose baselines using machine learning, neural networks, and transformer-based language models. The availability of a diverse dataset is a critical step towards democratizing NLP and ultimately contributing to a better understanding of code-mixed languages.

% We further evaluate the dataset on $11$ baseline methods, achieving an accuracy of $69.5\%$ and an F1 score of $68.8\%$.

\end{abstract}
% The dataset and code are available at: \url{https://github.com/Nishita2000/BnSentMix}. 
\section{Introduction}

\begin{figure}[ht]
  \includegraphics[width=1.07\columnwidth]{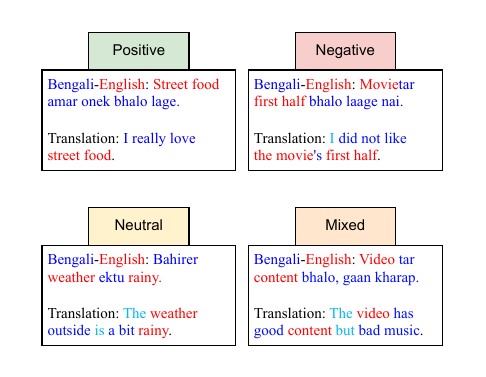}
  \caption{Examples of the four sentiment labels from our code-mixed Bengali-English dataset \textsc{BnSentMix} and the corresponding English translations. \textcolor{red}{Red} represents English words, \textcolor{blue}{blue} represents Bengali words written in English alphabets, and \textcolor{cyan}{cyan} represents implicit words in the code-mixed text.}
  \label{fig:sentimentExamples}
\end{figure}

\begin{table*}[ht!]
\centering
\resizebox{\textwidth}{!}{%
    \begin{tabular}{l c  c  c  c  c  c}
    \toprule
    \textbf{Dataset}  & \textbf{\#Samples} & \textbf{\#SL} & \textbf{\#DS} & \textbf{Filtering} & \textbf{\#Baselines} & \textbf{PA}
    \\ 
    \midrule
%      Hindi \cite{joshi-etal-2016-towards}  &  3.9k & 3& 1 & Manual &  10  & \textcolor{green}{\ding{51}} \\ % \hline
     
% Bengali \cite{Mandal2018PreparingBC} & 5k& 3 & 1 & Manual &  2  & \textcolor{red}{\ding{55}}\\ % \hline
     
% Tamil \cite{chakravarthi-etal-2020-corpus} &  15.7k & 5 & 1 & \texttt{langdetect} &  2  & \textcolor{green}{\ding{51}} \\ % \hline
% Malayalam \cite{chakravarthi-etal-2020-sentiment} &  6.7k & 5 & 1 & \texttt{langdetect} &  2 & \textcolor{green}{\ding{51}}\\ % \hline
% Persian \cite{sabri2021sentiment} &  3.6k & 3 & 1 & N/A &  2  & \textcolor{green}{\ding{51}} \\ % \hline
% Swiss \cite{pustulka2018multilingual} &  900& 3& 1 & N/A &  0  & \textcolor{green}{\ding{51}} \\ % \hline
% \textbf{BnSentMix (Ours)} &  \textbf{20k}& \textbf{4} & \textbf{3} & \textbf{mBERT} & \textbf{11}  & \textcolor{green}{\ding{51}} \\ \bottomrule
Hindi \cite{joshi-etal-2016-towards}  &  3.9k & 3& 1 & Manual &  10  & \textcolor{green}{\ding{51}} \\ % \hline
     
Bengali \cite{Mandal2018PreparingBC} & 5k& 3 & 1 & Manual &  5  & \textcolor{red}{\ding{55}}\\ % \hline
     
Tamil \cite{chakravarthi-etal-2020-corpus} &  15.7k & 5 & 1 & \texttt{langdetect} &  10  & \textcolor{green}{\ding{51}} \\ % \hline
Malayalam \cite{chakravarthi-etal-2020-sentiment} &  6.7k & 5 & 1 & \texttt{langdetect} &  10 & \textcolor{green}{\ding{51}}\\ % \hline
Persian \cite{sabri2021sentiment} &  3.6k & 3 & 1 & Keywords search &  3  & \textcolor{green}{\ding{51}} \\ % \hline
Swiss \cite{pustulka2018multilingual} &  963& 3& 1 & Manual &  7  & \textcolor{red}{\ding{55}} \\ % \hline
\textbf{BnSentMix (Ours)} &  \textbf{20k}& \textbf{4} & \textbf{3} & \textbf{mBERT} & \textbf{14}  & \textcolor{green}{\ding{51}} \\ \bottomrule
    \end{tabular}}
    \caption{Comparison of the number of samples, \#SL: Sentiment Labels, \#DS: Data Sources, filtering method, number of baselines, and PA: Public Availability of various code-mixed (with English) sentiment analysis datasets.}
    \label{tab:related_works_table}
\end{table*}

In the rapidly evolving digital landscape, code-mixing has become increasingly prevalent, particularly in multilingual societies. Code-mixing is the phenomenon of alternating between two or more languages within a single conversation or sentence \cite{thara2018code}. Code-mixing can occur in various forms, including intra-sentential switching, where words from different languages appear within the same sentence, and intra-word switching, where elements from other languages combine to form a single word \cite{stefanich2019morphophonology,litcofsky2017switching}. Intra-sentential switching is more frequently observed in colloquial settings. One significant yet understudied domain of code-switching is Bengali-English code-mixed text. 

We consider Fig. \ref{fig:sentimentExamples} where the sentences are examples of Bengali-English intra-sentential switching. Intra-word switching is observed in the negative sentiment example. Here, \textcolor{red}{Movie}\textcolor{blue}{tar} is considered a single word, whereas the Bengali sub-word \textcolor{blue}{tar} indicates possession. We also observe several words in the transliterated text that are not explicitly written in the code-mixed text. These implicitly defined words increase the challenges in processing code-mixed Bengali-English texts.

With over 250 million native speakers globally, Bengali is the seventh most spoken language in the world but remains a low-resource language in terms of research. While typing texts, Bengali speakers often use Bengali-English code-mixed terms to express their thoughts in writing. Despite the prevalence of code-mixed text on social media platforms, e-commerce sites, and other digital spaces, there remains a notable scarcity of resources to analyze and process such data.

Sentiment analysis, the computational study of people's opinions, sentiments, emotions, and attitudes expressed in written language, plays a critical role in various applications, including social media monitoring, customer feedback, market research, and public opinion analysis \cite{wankhade2022survey}. While substantial progress has been made in monolingual sentiment analysis \cite{medhat2014sentiment,birjali2021comprehensive}, the complexities introduced by code-mixed texts present unique challenges that current models struggle to address \cite{barman2014code}. This is particularly true for Bengali-English code-mixed texts \cite{chanda2016unraveling}, which have not received adequate attention in existing research.

Table \ref{tab:related_works_table} highlights the limitations of Bengali-English code-mixed sentiment analysis datasets compared to other Indic-English code-mixed datasets. The only available Bengali dataset \cite{Mandal2018PreparingBC} is limited to $5k$ samples, $3$ sentiment labels, a single data source, $5$ baselines, and is not publicly available. The existing language detection tools also have severe limitations in filtering code-mixed Bengali-English. Tools like \texttt{langdetect}\footnote{\url{https://pypi.org/project/langdetect/}} and 
Bengali phonetic parser\footnote{\url{https://github.com/porimol/bnbphoneticparser}} designed for general language identification and code-mixed Bengali identification struggled with the spelling nuances of code-mixed text. 

Addressing these challenges, our contribution can be summarized:
\begin{itemize}
    \item We present, \textsc{BnSentMix}, a novel Bengali-English code-mixed dataset comprising 20,000 samples and 4 sentiment labels for sentiment analysis. Data has been curated from YouTube, Facebook, and e-commerce platforms to encapsulate a broad spectrum of contexts and topics.
    
    \item Following the intricacies of code-mixed test, visualized in Fig. \ref{fig:sentimentExamples}, we propose a novel automated code-mixed text detection pipeline using fine-tuned language models, reaching an accuracy of 94.56\%.

    \item We establish 11 baselines including classical machine learning, neural network, and pre-trained transformer-based models, with BERT achieving accuracy and F1 score of 69.5\% and 68.8\% respectively.
\end{itemize}

\begin{figure*}[ht]
    \centering
  \includegraphics[width=0.9\textwidth]{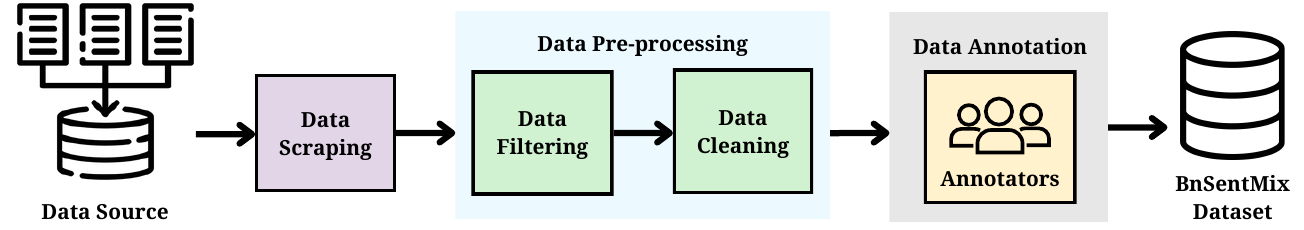}
  \caption{Dataset creation pipeline of the \textsc{BnSentMix} dataset.}
  \label{fig:datasetPipeline}
\end{figure*}

\section{Related Work}
\subsection{Code-Mixing} 
Code-mixed data can be the source of several text classification tasks \cite{thara2018code} with sentiment analysis \cite{mahadzir2021sentiment} being one of the most popular ones. Other natural language processing tasks (NLP) on code-mixed data include hate speech detection \cite{sreelakshmi2020detection}, translation \cite{gautam2021comet}, part of speech tagging \cite{vyas2014pos}, emotion classification \cite{ameer2022multi}, language identification \cite{mandal2018language}, and speech synthesis \cite{sitaram2016speech}. Researchers also incorporate training data augmentation \cite{gupta2021training,rizvi2021gcm} and code-mix word embeddings \cite{pratapa2018word} to process code-mixed texts.

\subsection{Sentiment Analysis}
The significance of sentiment analysis has grown with the rise of social media, prompting extensive research on monolingual corpora. Studies explored various languages, including English \cite{hu2004mining, wiebe2005annotating, jiang2019challenges}, Russian \cite{rogers2018rusentiment}, German \cite{cieliebak2017germansentiment}, Norwegian \cite{maehlum2019sentiment}, several Indian languages \cite{agrawal2018deep,rani2020sentiment}, and Bengali \cite{fahim-2023-aambela-blp, kabir-etal-2023-banglabook}. Multilingual sentiment analysis \cite{dashtipour2016multilingual,pustulka2018multilingual} gained popularity with the recent advancements in multilingual language models \cite{devlin-etal-2019-bert,conneau-etal-2020-unsupervised}. 

% Over the years, a vast number of code-mixing datasets with English have been proposed, as enlisted in Table \ref{tab:related_works_table}.

% Chakravarthi et al. developed a gold standard corpus for sentiment analysis of code-mixed Malayalam-English text from 6,739 annotated YouTube comments, demonstrating high inter-annotator agreement and strong performance from traditional machine learning and deep learning models, except for SVM which struggled with the non-Malayalam class \cite{chakravarthi-etal-2020-sentiment}. Joshi et al. proposed a sentiment analysis approach for Hindi-English code-mixed text using Subword-LSTM on a dataset of 4,981 sentiment-annotated Facebook comments, outperforming character-level and word-level models by 18\% in accuracy \cite{joshi-etal-2016-towards}. Chakravarthi et al. also created a Tamil-English code-mixed sentiment analysis corpus from 15,744 YouTube comments, providing benchmarks with various machine learning models, including SVM, Decision Tree, and BERT-Multilingual \cite{chakravarthi-etal-2020-corpus}. 

\subsection{Code-Mixing in Bengali}

Bengali is often code-mixed with English \cite{chanda2016unraveling} and Hindi \cite{raihan2023sentmix}. In Bengali-English code-mixing, English tokens are commonly used alongside romanized or transliterated Bengali \cite{shibli2023automatic, fahim-etal-2024-banglatlit}, which is often back-transliterated before processing \cite{haider2024banth}. Sentiment analysis on code-mixed Bengali has limited studies, either using small private datasets \cite{Mandal2018PreparingBC} or performed in a multilingual setting \cite{patra2018sentiment}. Data augmentation techniques have also been explored to enhance code-mixed sentiment analysis datasets in Bengali \cite{tareq2023data}. Emotion detection, a task similar to sentiment analysis, has also been studied in the context of code-mixed Bengali \cite{raihan2024emomix}.

\section{\textsc{BnSentMix} Dataset}
The \textsc{BnSentMix} data has been collected from multiple data sources to reflect realistic code-mixed texts commonly found in digital spaces. We labeled the data using four distinct sentiments: the commonly used positive, negative, and neutral sentiments, as well as a \emph{mixed} sentiment. As illustrated in Fig. \ref{fig:sentimentExamples}, the mixed sentiment represents instances where both positive and negative sentiments are conveyed within different parts of the text. We decided to include the mixed label because the associated sentences are frequently observed in everyday texts and cannot be correctly classified under the traditional sentiment labels.

\begin{figure}[ht]
  \includegraphics[width=\columnwidth]{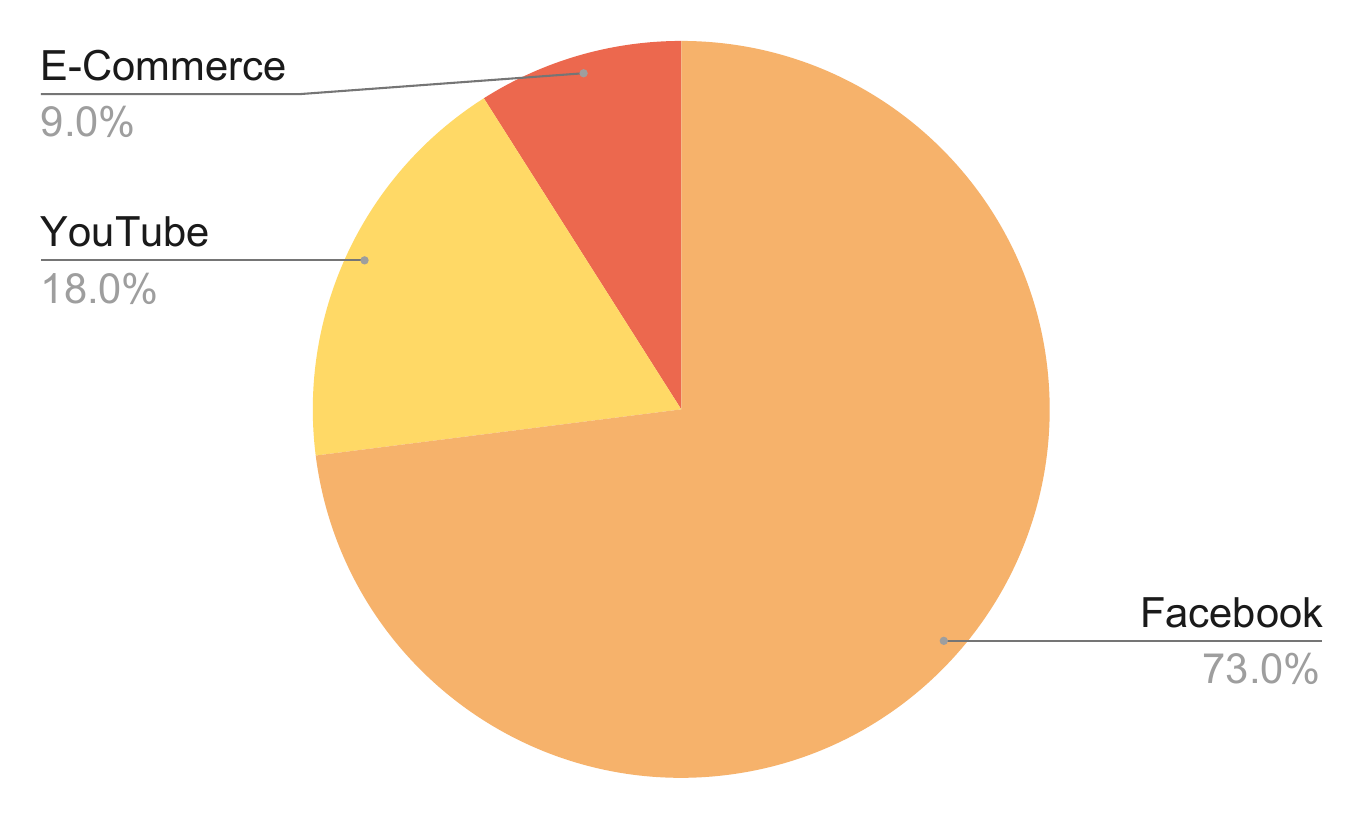}
  \caption{Composition of data sources of the \textsc{BnSentMix} dataset.}
  \label{fig:scraping}
\end{figure}

\subsection{Data Sourcing}
\label{sec:dataSourcing}
We collected extensive user-generated content from YouTube comments, Facebook comments, and e-commerce site reviews. These data sources were chosen for their high engagement rates and diverse linguistic input. YouTube comments were scraped using the YouTube API. We used \texttt{Facepager}\footnote{\url{https://github.com/strohne/Facepager}} to extract comments from public Facebook posts, pages, and groups. \texttt{Selenium}\footnote{\url{https://selenium-python.readthedocs.io/}} was employed to mimic human browsing behavior on e-commerce sites to scrape product reviews. We amassed over 3 million samples of user-generated content, forming the foundation for our dataset and subsequent analysis. Fig. \ref{fig:scraping} illustrates the composition of the aforementioned data sources.

\subsection{Data Cleaning}

% \begin{figure*}[ht]
%   \includegraphics[width=\textwidth,height= 4cm]{methodology2.pdf}
%   \caption{Methodology for Data Filtering}
%   \label{fig:dataProcessing}
% \end{figure*}

We discard samples with four words or less and samples containing external URLs. Redundant whitespaces, special characters, and non-ASCII characters including emojis and emoticons are also removed. Consequent sequences of punctuation symbols are reduced to single instances. The English words are downcased unless they appear at the beginning of the sentence. However, we did not correct any form of typing or grammatical errors in our dataset to ensure the trained model is robust for practical scenarios. The data cleaning procedure has been formally described in Algo. \ref{algo:dataClean}.
\begin{algorithm}[ht]

\caption{Clean Text}
\label{algo:dataClean}
\begin{algorithmic}[1]
\REQUIRE $text \gets$ Input text
\ENSURE $text \gets$ Preprocessed text

\STATE $text \gets text.lower()$ \COMMENT{Convert to lowercase}
\STATE $text \gets$ Remove all special characters except "?", ",", "!", and "."
\STATE $text \gets$ Reduce consecutive sequences of punctuations to a single instance
\STATE $text \gets$ Remove all non-ASCII characters
\STATE $text \gets$ Remove extra white spaces
\STATE $text \gets$ Capitalize the first letter after each period (.)

\RETURN $text$
\end{algorithmic}
\end{algorithm}

\subsection{Data Filtering}

We construct a novel Bengali-English code-mix detection dataset and fine-tune pre-trained language models to automatically filter code-mixed Bengali-English. Detecting these texts can pose significant challenges: (i) rule-based methods struggle with intra-word switching (ii) romanized Bengali or English samples may be incorrectly classified as code-mixed text by automated methods, and (iii) samples from a third language often bypass the filtering process. Our approach addresses these challenges by incorporating pre-trained language models, which excel in intricate text detection settings. Algo. \ref{algo:dataFiltering} outlines the data filtering pipeline. 

\begin{algorithm}[ht]
\caption{Detect Code-mixed Bengali}
\label{algo:dataFiltering}
\begin{algorithmic}[1]
\REQUIRE $S \gets$ List of sentences
\REQUIRE $model \gets$ Pre-trained mBERT model
\REQUIRE $tokenizer \gets$ Pre-trained mBERT tokenizer
\ENSURE $pred \gets$ Predicted class label (0 or 1)
\STATE $b\_count \gets 0$
\STATE $w\_count \gets 0$
\FOR{each $sent$ in $S$}
    \STATE $words \gets$ split($sent$)
    \FOR{each $w$ in $words$}
        \STATE $w \gets$ preprocess($w$)
        \IF{$w$ is empty}
            \STATE continue
        \ENDIF
        \STATE $w\_count \gets w\_count + 1$
        \STATE $inputs \gets$ tokenize($w$)
        \STATE $outputs \gets$ model($inputs$)
        \STATE $pred\_class \gets$ argmax($outputs$)
        \IF{$pred\_class == 1$}
            \STATE $b\_count \gets b\_count + 1$
        \ENDIF
    \ENDFOR
\ENDFOR
\IF{$w\_count < 4$}
    \RETURN 0
\ENDIF
\STATE $b\_percent \gets b\_count / w\_count$
\IF{$b\_percent \geq 0.3$}
    \RETURN 1
\ELSE
    \RETURN 0
\ENDIF
\end{algorithmic}
\end{algorithm}

\subsubsection{Code-mix Detection Dataset}
% \begin{figure}[ht]
%   \includegraphics[width=\columnwidth]{first_dataset.pdf}
%   \caption{Banglish-English Mixed Dataset}
%   \label{fig:first_dataset}
% \end{figure}
% To create a robust dataset for pre-training, we sourced and combined three key datasets to ensure comprehensive coverage of both Bengali and English. The first dataset was Google's Dakshina dataset, a rich collection of text in South Asian languages, including many Benglish (Bengali-English code-mixed) sentences, which were instrumental for our purposes. The second source was a Kaggle English dataset, from which we collected a wide range of English words. For the third source, we collected more data from Mandal et al. By integrating these diverse sources, we curated a comprehensive dataset of 100,000 words, ensuring a balanced mix of Bengali, English, and code-mixed text. This dataset provided a solid foundation for fine-tuning our pre-trained models.

The fine-tuning dataset for code-mixed Bengali-English detection comprises $3$ data sources. We incorporate the Dakshina dataset \cite{roark-etal-2020-processing} which has a rich collection of Southeast Asian languages, including many Bengali-English code-mixed sentences. Secondly, we utilized a Kaggle English dataset\footnote{https://www.kaggle.com/datasets/rtatman/english-word-frequency} consisting of a wide range of English words and extended with a third source \citet{mandal2018language}. By integrating these diverse sources, we curated a comprehensive dataset of $100k$ words, ensuring a balanced mix of Bengali, English, and code-mixed Bengali-English words. To maintain the linguistic purity of code-mixed Bengali-English, we exclude sentences containing words that are neither English nor Bengali, e.g. Hindi words.

\subsubsection{Code-mix Detection Results}
We evaluate $3$ pre-trained models -- the multilingual models, mBERT \cite{devlin-etal-2019-bert} and XLM-RoBERTa \cite{conneau-etal-2020-unsupervised}, and the Bengali-English model BanglishBERT \cite{bhattacharjee-etal-2022-banglabert}. Table \ref{tab:pre_results} reveals mBERT showing substantially higher accuracy and F1 score in code-mixed Bengali-English detection. We argue that the pre-trained multilingual capabilities of mBERT effectively handled the nuances of code-mixed Bengali-English text. 

\begin{table}[ht]
\begin{center}
    \begin{tabular}{lcc}
    \toprule
    \textbf{Model} & \textbf{Acc(\%)} & \textbf{F1 Score}\\
    \midrule  
    XLM-RoBERTa & 89.60 & 0.8985 \\
    BanglishBERT & 90.56 & 0.8961 \\
    \textbf{mBERT} & \textbf{94.56} & \textbf{0.9403} \\
     \bottomrule
    \end{tabular}
\end{center}
    \caption{Comparison of the accuracy and F1 score of the code-mixed Bengali-English detection methods.}
    \label{tab:pre_results}
\end{table}

% \subsubsection{Filtering Pipeline}
% Following Fig. \ref{fig:datasetPipeline}, the code-mixed Bengali-English is cleaned and passed to the data filtering classifier. 
% % The filtered code-mixed Bengali is manually verified to ensure the highest quality of data. To maintain linguistic purity, we exclude sentences containing words that are not English or Bengali. 
% We discard samples containing hate speech and offensive content. We also discarded samples with a small proportion of code-mixing. Sentences are considered to be code-mixed Bengali-English if they contain at least 30\% English words. We obtain 21,587 data samples after data filtering.

\begin{figure}[ht]
  \includegraphics[width=\columnwidth]{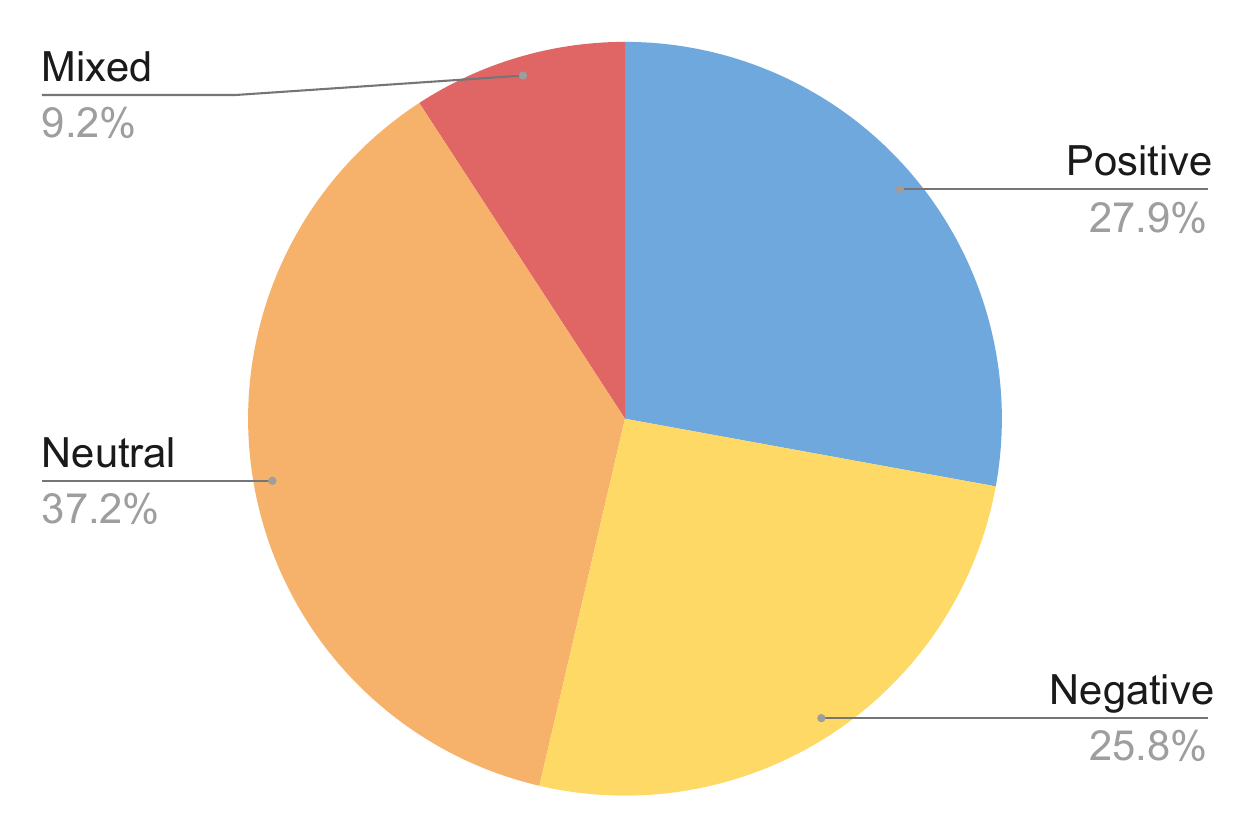}
  \caption{Distribution of sentiment labels in the \textsc{BnSentMix} dataset.}
  \label{fig:labelPieChart}
\end{figure}

\subsection{Data Annotation}

Each sample in our dataset has been annotated twice by two different annotators to ensure generalized sentiment is conveyed. In cases where the two independent annotations did not match, a third annotator would break the tie. To perform data annotation, we recruited $64$ annotators who had been provided hourly monetary compensation. The data annotators have at least a high-school degree (equivalent to Grade 12 education) and are familiar with social media and digital spaces. The annotators were asked to re-label the same $250$ samples to measure inter-annotator agreement. We measured the agreement score using Cohen's Kappa $\kappa= 0.86$, indicating substantial agreement. 

\subsection{Dataset Statistics}
% After a rigorous evaluation process, 20,000 samples were finalized from the 21,587 samples, ensuring high-quality annotations for subsequent analysis and model training. 

Fig. \ref{fig:labelPieChart} visualizes the label composition of the annotated dataset. An overview of the key statistics of the annotated dataset is shown in table-\ref{tab:dataStats}. We split the dataset into $[70:15:15]$ training, validation, and test splits i.e. 14,000, 3,000, and 3,000 samples respectively.

\section{Methodology and Experimental Setup}

\subsection{Baseline Models}
We evaluate $11$ baselines encompassing traditional machine learning models, recurrent neural network variants, and transformer-based pre-trained language models, observed in table \ref{tab:mainPerformanceTable}. All the pre-trained models were fine-tuned on our dataset.
% with BERT \cite{devlin-etal-2019-bert} achieving the best performance of 69.5\% test accuracy and 68.8\% test F1 score.

% \subsection{Further Pre-trained Transformer Models}
% \label{sec:furtherPretraining}
% We enhance the performance of the transformer baselines by incorporating Masked Language Modeling loss \cite{devlin-etal-2019-bert} to further pre-train the models on the unlabeled code-mixed data corpus scraped in sec-\ref{sec:dataSourcing}. The MLM loss following dynamic masking can be formulated as:
% \begin{equation}
%     L_{\text{MLM}} = - \frac{1}{N} \sum_{i=1}^{N} \log P(x_i \mid \mathbf{x}_{\text{mask}})
% \end{equation}
% where, \( x_i \) is the token at position \( i \) masked in the input sequence, \( \mathbf{x}_{\text{mask}} \) denotes the input sequence with dynamically masked tokens, \( N \) is the total number of masked tokens. We ensure that samples from the test and validation sets are excluded from the further pre-training corpus. The further pre-trained models are referred to as Code-Mixed Bengali (CMB) transformer models and they follow the naming convention $x$-CMB, where $x$ represents a pre-trained transformer variant.

\subsection{Evaluation Metrics}

We use classification accuracy and F1-score for model evaluation -- both well-known metrics for text classification \cite{hossin2015review}.

\begin{table}[ht]
\begin{center}
    \begin{tabular}{lr}
    \toprule
    \textbf{Statistic} & \textbf{Value}\\
    \midrule  
  Mean Character Length & 62.77 \\
Max Character Length & 1985 \\
Min Character Length & 14 \\
Mean Word Count & 11.65 \\
Max Word Count & 368 \\
Min Word Count & 4 \\
Unique Word Count & 37734 \\
Unique Sentence Count & 20000 \\
     \bottomrule
    \end{tabular}
\end{center}
    \caption{Key statistics of the \textsc{BnSentMix} dataset.}
    \label{tab:dataStats}
\end{table}
\subsection{Implementation Details}
The models were trained on NVIDIA Tesla P100 GPUs with 16GB of memory. We followed the Huggingface implementation \cite{wolf2019huggingface} for the pre-trained language models. All the models utilized Adam Optimizer \cite{kingma2014adam} with a training batch size of $32$. The training configuration used most of the default hyperparameters. Logistic Regression, RNN, and LSTM models used the learning rate of $1E{-5}$ while the BERT-family language models used the learning rate of $1.5E{-6}$. The training time for each epoch varied from $8$ to $13$ minutes.
 
\begin{table*}[ht]
\begin{center}
    \begin{tabular}{l cccc cccc}
    \toprule

\multirow{2}{*}{\textbf{Model}} & \multicolumn{4}{c}{\textbf{Validation}} & \multicolumn{4}{c}{\textbf{Test}} \\ \cmidrule(lr){2-5}\cmidrule(lr){6-9}
 & \textbf{Acc}           &  \textbf{Precision}         &     \textbf{Recall}       &    \textbf{F1} & \textbf{Acc}           &  \textbf{Precision}         &     \textbf{Recall}       &    \textbf{F1} \\   \midrule
\multicolumn{9}{l}{\textbf{Machine Learning Models}} \\
  ~~~~~Logistic Regression    & 0.668                   & 0.656                & 0.668                & 0.662                & 0.667                & 0.614                & 0.667                & 0.639                \\
~~~~~Random Forest          & 0.672                   & 0.661                & 0.672                & 0.666                & 0.648                & 0.635                & 0.648                & 0.641                \\
~~~~~SVM                    & 0.694                   & 0.676                & 0.694                & 0.685                & 0.660                & 0.637                & 0.660                & 0.648                \\ \midrule
\multicolumn{9}{l}{\textbf{Recurrent Neural Network Variants}} \\
~~~~~RNN                    & 0.406                   & 0.308                & 0.406                & 0.350                & 0.401                & 0.352                & 0.401                & 0.375                \\
~~~~~LSTM                   & 0.678                   & 0.670                & 0.678                & 0.674                & 0.670                & 0.657                & 0.670                & 0.663                \\ \midrule
\multicolumn{9}{l}{\textbf{Multilingual Language Models}} \\
~~~~~XLM-RoBERTa            & 0.726                   & 0.709                & 0.726                & 0.717                & 0.698                & 0.642                & 0.698                & 0.669                \\
~~~~~mBERT                  & 0.726                   & 0.713                & 0.726                & 0.719                & 0.694                & 0.675                & 0.694                & 0.684                \\\midrule
\multicolumn{9}{l}{\textbf{Bangla Language Models}} \\
~~~~~BanglaBERT             & 0.721                   & 0.668                & 0.721                & 0.693                & 0.698                & 0.642                & 0.698                & 0.669                \\
~~~~~BanglishBERT           & 0.694                   & 0.715                & 0.694                & 0.704                & 0.686                & 0.653                & 0.686                & 0.669                \\\midrule
\multicolumn{9}{l}{\textbf{English Language Models}} \\
~~~~~DistilBERT             & 0.701                   & 0.694                & 0.701                & 0.697                & 0.672                & 0.665                & 0.672                & 0.668                \\
~~~~~BERT          & 0.727          & 0.710       & 0.724       & 0.717       & 0.695       & 0.683       & 0.694       & 0.688    \\
%  \midrule
% \multicolumn{9}{c}{Code-Mixed Bengali (CMB) Transformer Models} \\
%  \midrule
%  mBERT-CMB                  & 0.727                   & 0.713                & 0.727                & 0.720                & 0.694                & 0.675                & 0.695                & 0.685                \\
% BanglaBERT-CMB             & 0.723                   & 0.669                & 0.724                & 0.695                & 0.699                & 0.643                & 0.699                & 0.670                \\
 
%  \textbf{BERT-CMB}          & \textbf{0.730}          & \textbf{0.712}       & \textbf{0.728}       & \textbf{0.720}       & \textbf{0.698}       & \textbf{0.685}       & \textbf{0.698}       & \textbf{0.691}    \\
     \bottomrule
    \end{tabular}
\end{center}
    \caption{Performance of the proposed baselines based on accuracy, precision, recall, and F1 score.}
    \label{tab:mainPerformanceTable}
\end{table*}

\section{Results and Analysis}

\subsection{Performance Evaluation}
Table \ref{tab:mainPerformanceTable} highlights the performance of the $11$ baselines with BERT achieving the best performance in terms of both accuracy and F1 score. We now analyze the category-wise model performance.

 \subsubsection{Machine Learning (ML) Models}
The ML models provide simple baselines and achieve considerably high accuracy, with the Support Vector Machine (SVM) \cite{vapnik95} achieving accuracy and F1 score on par with larger transformer-based models like BanglishBERT. The other two ML baselines Logistic Regression \cite{cox1958regression} and Random Forest \cite{breiman2001random} achieve satisfactory performance with relatively simpler architectures. These ML baselines can be effective in resource-constrained scenarios.

\subsubsection{Recurrent Neural Networks (RNNs)}
RNN \cite{hopfield1982neural} underperformed compared to the other baselines. On the contrary, the performance of Long Short-Term Memory (LSTM) models \cite{hochreiter1997long} was significantly higher in terms of both accuracy and F1 score. We argue that the long-term textual dependencies and the impact of vanishing and exploding gradients limited the performance of the RNN models.

\begin{figure}[ht]
  \includegraphics[width=\columnwidth]{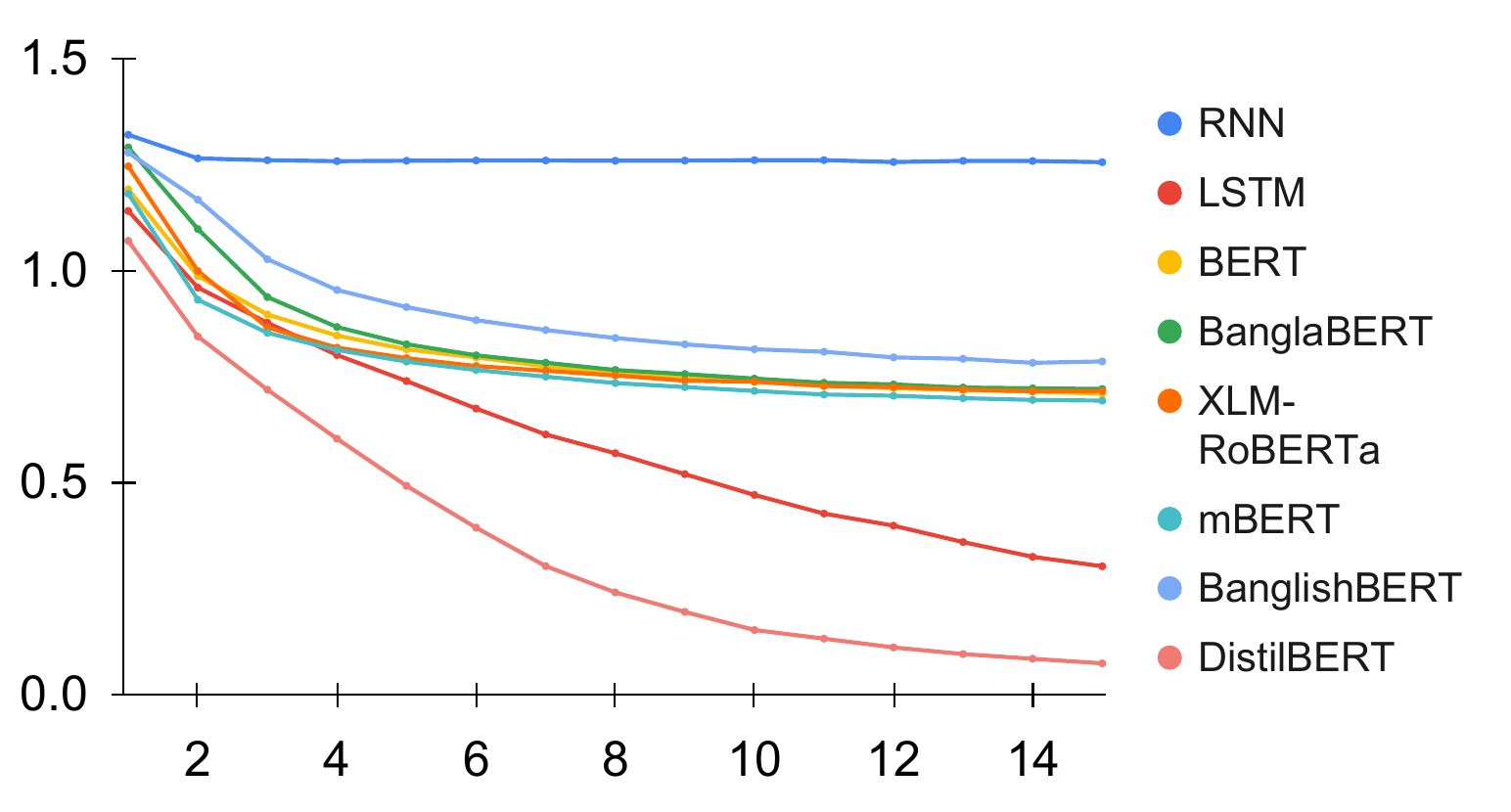}
  \caption{Comparison of epoch-wise training loss of the established baselines.}
  \label{fig:training_loss}
\end{figure}

\subsubsection{Transformer-based Models}

 The best performance is achieved by the BERT model \cite{devlin-etal-2019-bert} pre-trained on an English corpus. The BERT model is closely followed by the multilingual models XLM-RoBERTa \cite{conneau-etal-2020-unsupervised} and mBERT \cite{devlin-etal-2019-bert}. We hypothesize that the low proportion of Bengali text in the multilingual pre-training corpus does not provide any significant advantage in code-mixed Bengali classification tasks. 
 
In contrast, English pre-trained models like BERT exhibit better understanding of the linguistic intricacies of English words used in code-mixed Bengali, thereby producing better performance than other multilingual and Bengali models. Similarly, the Bengali language models BanglaBERT \cite{bhattacharjee-etal-2022-banglabert} and BanglishBERT \cite{bhattacharjee-etal-2022-banglabert} are trained on Bengali and Bengali-English corpora respectively. Code-mixed Bengali uses English tokens and hence, the pre-training on Bengali tokens does not provide any significant advantage. The lighter version of BERT, DistilBERT \cite{sanh2019distilbert} produces comparable but slightly worse results.

% \subsubsection{Code-Mixed Bengali Models}
% We selected the three best-performing transformer encoders and further pre-trained them following sec-\ref{sec:furtherPretraining}. The models were selected from each sub-category -- mBERT as the best multilingual model, BanglaBERT as the best Bengali model, and BERT as the best English model. We find that further pre-training on the code-mixed data translates to an increase in performance for all three models. The only exception is the mBERT-CMB model showing insignificant improvement in accuracy and F1-score. Similar to the pre-trained baselines, the BERT-CMB is the best-performing model followed by mBERT-CMB and BanglaBERT-CMB. We argue that further pre-training in a larger code-mixed corpus might translate to better performance for multilingual models like mBERT or XLM-RoBERTa.   

\subsection{Training Loss Analysis}

Figure~\ref{fig:training_loss} illustrates the training loss across $15$ epochs for the baselines. We observe that all models converge before reaching the $15^{th}$ epoch. The only exception is the LSTM model which shows a slight indication of being benefited by additional training epochs. Excluding DistilBERT, the other BERT family models converged relatively faster in the earlier epochs. For most models, training for $5$-$8$ epochs is appropriate to prevent overfitting. 

\section{Conclusion}
We introduce \textsc{BnSentMix}, a novel sentiment analysis dataset tailored for code-mixed Bengali-English. Our work opens several potential research avenues for code-mixed Bengali. Researchers can explore other tasks, such as hate speech, offensive language, and abusive content detection on code-mixed data. Our work addresses a significant gap for low-resource languages and sets a new standard for sentiment analysis in code-mixed Bengali-English.

\section*{Data Availability}
Our dataset will be publicly available under the Creative Commons Attribution 4.0 International (CC BY 4.0). Any form of private data or personal identification information has been removed from the dataset to prevent privacy violations. We have ensured that the redistribution of social media data is consistent with the policies of the corresponding platforms.

\section*{Limitations}
The label distribution of \textsc{BnSentMix} dataset is slightly imbalanced with only 9.2\% samples labeled as mixed sentiment which can affect the performance of the model in classifying mixed sentiments. We also acknowledge that the sentiment of the annotator can be a source of bias during data annotation, though each data sample has been annotated twice by two different annotators, and annotation conflicts have been resolved by a third annotator. 

% Emojis and emoticons can convey sentiment but have been removed from our dataset. Their impact on the code-mixed sentiment can be explored in future works. Finally, complex sentiments involving sarcasm and humor are usually classified as neutral but can be subjective depending on the context.

\section*{Ethical Statement}

The hired data annotators were compensated significantly higher than the region's minimum wage. Each annotator was only given around $630$ data samples with no time restrictions. This ensured that the annotator did not overwork during data annotation. Annotator sentiment is subject to long working hours and can affect sentiment labeling. To prevent this, we mandated five-minute breaks after every twenty-minute interval and provided refreshments upon request.

\section*{Acknowledgements}
Our work is supported by the Islamic University of Technology Research Seed Grants (IUT RSG) (Ref: REASP/IUT-RSG/2022/OL/07/012). We sincerely appreciate Mohammed Saidul Islam and Md Mezbaur Rahman for guidance and Nejd Khadija for proofreading our work.
\bibliography{custom}

\end{document}